\documentclass{article}

\usepackage{PRIMEarxiv}

\usepackage[utf8]{inputenc} 
\usepackage[T1]{fontenc}    
\usepackage{hyperref}       
\usepackage{url}            
\usepackage{booktabs}       
\usepackage{amsfonts}       
\usepackage{nicefrac}       
\usepackage{microtype}      
\usepackage{lipsum}
\usepackage{fancyhdr}       
\usepackage{graphicx}       
\usepackage{float}
\graphicspath{{images/}}     
\usepackage{mathrsfs,amsmath}   
\usepackage[frozencache,cachedir=.]{minted}
\usepackage{listings}
\title{Scaling Up Computer Vision Neural Networks using Fast Fourier Transform
\thanks{\textit{\underline{Citation}}: 
\textbf{Authors. Title. Pages.... DOI:000000/11111.}} 
}

\author{
  Siddharth Agrawal \\ Ahmedabad University
}

\begin{document}
\maketitle

\begin{abstract}
Deep Learning-based Computer Vision field has recently been trying to explore larger kernels for convolution to effectively scale up Convolutional Neural Networks. Simultaneously, new paradigm of models such as Vision Transformers find it difficult to scale up to larger higher resolution images due to their quadratic complexity in terms of input sequence. In this report, Fast Fourier Transform is utilised in various ways to provide some solutions to these issues.
\end{abstract}


\section{Introduction}
While Fourier Transform has seen many applications in data compression and signal processing, it's applications in Deep Neural Networks are limited. They are often only used for Medical Imaging based applications. Here, I discuss three approaches of scaling up Neural Networks for computer vision using Fast Fourier Transform (FFT).

Note: python libraries were used for FFT as they provide extremely efficient cuda-based approaches for FFT on the GPU. The report only contains part of the code; the entire code-base is very large and includes the dataloaders, hyperparameter configurations, scripts to test fps, datasets, training and validation engines, and the model implementations themselves. The entire codebase can be found at \href{https://github.com/siddagra/fourier-project}{https://github.com/siddagra/fourier-project}

\section{Fourier Vision Transformers}
\subsection{Introduction to Vision Transformers}
Vision Transformers (ViT) \cite{ViT} extended the use of transformers to the domain of computer vision. Typically, the quadratic $O(n^2)$ complexity in terms of input causes transformers like the one proposed in the original work \cite{transformers} for Natural Language Processing (NLP) to perform poorly in high dimensional tasks such as computer vision, where input size is large. ViT proposed to scale transformers up to images by deviding the image into 16x16 patches and computing patch embeddings of these.

Given an image tensor $I \in R^{W\times H \times C}$, patch embeddings are computed via a $16x16$ convolution with $n$ number of kernels. Convolution stride is set to patch size. The final output is flattened and transposed along the last two dimensions, resulting in embeddings tensor $P \in R^{\frac{p\times H}{16}}$. Lastly, learned embeddings of the patches position within the image are added to each patch embedding.

The process is shown in the code below:
\begin{minted}{python}
class PatchEmbeddings(nn.Module):
    def __init__(self, config):
        super().__init__()
        img_size = config.img_size
        patch_size = config.patch_size
        grid_size = (img_size[0] // patch_size[0],
                     img_size[1] // patch_size[1])
        num_patches = grid_size[0] * grid_size[1]
        self.cls_token = nn.Parameter(torch.randn(1, 1, config.embed_dim))

        self.proj = nn.Conv2d(
            config.in_chans, config.embed_dim, kernel_size=patch_size, stride=patch_size, bias=True)
        self.norm = nn.LayerNorm(config.embed_dim)
        self.positional_embeddings = nn.Parameter(
            torch.zeros(1, num_patches+1, config.embed_dim))

    def forward(self, x):
        B, C, H, W = x.shape
        x = self.proj(x)
        cls_token = self.cls_token.repeat(B, 1, 1)
        x = x.flatten(2).transpose(1, 2)
        x = torch.cat([cls_token, x], dim=1)
        x = x + self.positional_embeddings
        x = self.norm(x)
        return x  # B, C, P + 1
\end{minted}

The typical Vision Transformer would now use the multi-headed self-attention mechanism. Which can be summarised as the following operation:

$$
\begin{array}{c}
Attention(Q, K, V) = softmax(\frac{QK^T}{\sqrt{d_k}})V \\
head_i = Attention(Q W^Q_i, K W^K_i, V W^V_i) \\
MultiHead(Q, K, V) = Concat(head_1, ..., head_h)W^O \\
\end{array}
$$

where $W^K_i$, $W^Q_i$, $W^V_i$ and $W^O$ are matrices of learnable parameters, dictating the projection of the input.

Thus, in lay man terms, by the above formulation, the self-attention can look at the inputs to its layer, and decide how much importance or weightage to give all other inputs, when pondering over a specific token (image patch embeddings in this case) via the alignment between $Q$ and $K$ and then multiplying it with $V$ after normalisation and $softmax$. The multi-handedness allows the transformer to compute multiple such attention maps, and thus promotes thinking from different perspectives. Often, multiple different types of features extraction methods are required, and both global and local receptive field is required to achieve good results. Multi-headedness promotes this.

\begin{figure}[htp]
    \centering
    \includegraphics[width=\linewidth]{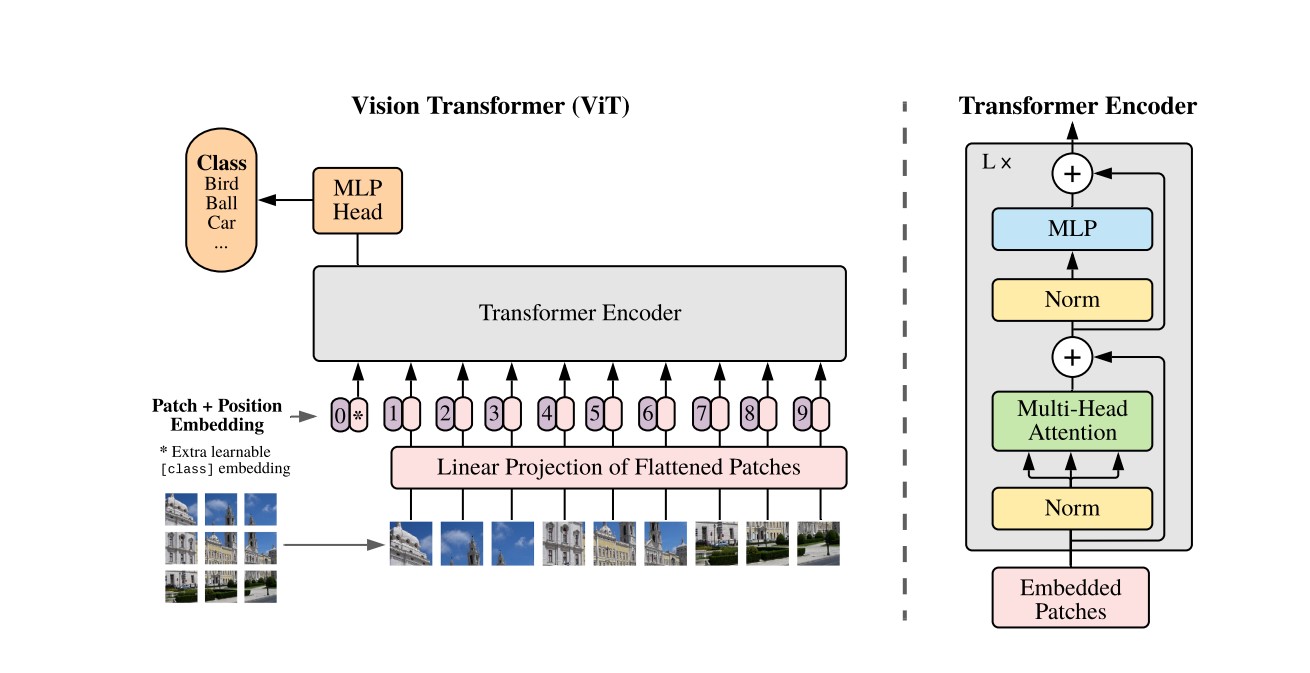}
    \caption{Overview of ViT. Images are first divided into 16x16 patches and the resulting tensor is flattened along its last two dimensions. Patch embeddings are computed for each patch and positional embeddings are added. The resulting tensor is fed into a vanilla transformer. Norm represents Layer Norm and MLP is a Multi Layer Perception. + denotes skip connections used to propagate the identity function.}
    \label{fig:galaxy}
\end{figure}

Specifically, the self-attention part comes from the fact that all: query ($Q$), key ($K$), and value ($V$) are equal to the input of the layer.

Layer Norm \cite{ba_kiros_hinton_2016} (Fig. 1) is normalisation applied before computations involving learnable parameters, as normalisation has shown to make learning easier and converge faster. This happens as it eliminates internal covariate shift involved with the $ReLU$ or $GeLU$ activation functions, and also brings all features to a similar scale \cite{ioffe_szegedy_2015}. The normalisation is computed using two parameters, $\epsilon$, $\gamma$ and $\beta$ and is summarised as follows:

$$
y=\frac{x-\mathrm{E}[x]}{\sqrt{\operatorname{Var}[x]+\epsilon}} * \gamma+\beta
$$

Thus, it is analogous to the $\gamma$ controlling for the vairance and the $\beta$ controlling the mean of the input feature's distributions. $\epsilon$ adds noise to the input, acting as a regulariser for the model.

A multilayer perceptron (MLP) (Fig. 1) is simply a recursive formulation wherein the input to the layer is linearly projected, and then an activation function is applied. This is repeated for some $n$ number of layers set by the developer.
Specifically, a single layer of MLP can be mathematically denoted as follows:

$$
y = act(W^T*x + b)
$$
where $act$ is some non-linear activation function such as $ReLU$, $GeLU$, $Sigmoid$, etc.

This transformer block is stacked multiple times, achieving a deep neural network with hierarchical representations. At the end, the output of a special  \lstinline|CLS| token is fed through an MLP to get the image classification.

\subsection{Proposed Approach: Fourier Image Transformers}
Due to the reliance of Vision Transformers on self-attention, it has a space and time complexity of $O(n^2)$ making it difficult to scale up to large images and higher resolutions. 

Thus, I borrow key insights from FNet \cite{FNet} which uses Fast Fourier Transform (FFT) to mix input signals in word embeddings. FNet was originally used in the domain of NLP, where it achieved then state-of-the-art results in long range arena benchmarks \cite{tay_dehghani_abnar_shen_bahri_pham_rao_yang_ruder_metzler_2020}. I try to extend this to the computer vision domain similar to \cite{ViT} via patch embeddings. FNet replaces the quadratic complexity self-attention layer, with an FFT operation to mix all tokens (embeddings) in $O(nlogn)$.

\begin{figure}[htp]
    \centering
    \includegraphics[width=\linewidth/3]{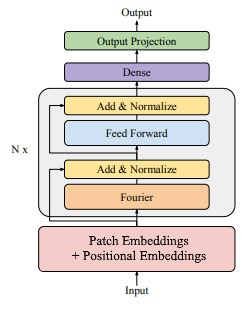}
    \caption{Overview of proposed architecture. Image and patch embeddings are computed as in ViT. However, the model now uses Fast Fourier Transform to mix the embeddings.}
    \label{fig:galaxy}
\end{figure}

The discrete Fourier Transform (DFT) is defined by the formula:
$$
X_k=\sum_{n=0}^{N-1} x_n e^{-\frac{2 \pi i}{N} n k}, \quad 0 \leq k \leq N-1
$$
and can be computed in $O(nlogn)$ using Fast Fourier Transform.

Similar to ViT, patch embeddings are generated for the image and flattened. However, now they are fed through a Fourier transform layer instead of a multi-headed self-attention layer.

Specifically, the layer applies a 2D DFT: a 1D DFT along the dimension of the sequence of all the patch embeddings ($\mathcal{F}_{seq}$), and another 1D DFT along the hidden/embedding dimension ($\mathcal{F}_{h}$). Empirically, it was found that extracting real part of the FFT at the end of the entire operation yielded best results. i.e., Real part was only extracted after applying both 1D FFTs: $\mathcal{F}_{h}$ and $\mathcal{F}_{seq}$.

The process can be summarised as follows:
$$
y = \mathcal{R}(\mathcal{F}_{h}(\mathcal{F}_{seq}(x))
$$

The rest of the model stays largely the same. Layer Norm is used after each Fourier and Feed Forward layer, and the identity function is propogated through the network via skip connections. A feed forward layer provides projection similar to $W^O$ from the vanilla transformer, linearly projecting the input to it.

These blocks are stacked multiple times to get a deeper model.

The model provides an efficient method of mixing embeddings with no additional parameter requirements in the mixing layer, and a much more scalable time complexity of $O(nlogn)$. Due to the duality of Fourier transform, each transformer block can be considered to alternatively apply Fourier transform and inverse Fourier transform, alternating between the spatial and the frequency domain, and each feed forward layer can be considered as a convolution (when in the frequency domain), and multiplications (when in the spatial domain). The negative sign introduced during the inverse Fourier transform can be inverted by the learnable feed forward layer, if it learns and finds it beneficial to do so.

\begin{minted}{python}
## Fourier Vision Transformer Block

class FFTLayer(nn.Module):
    def __init__(self):
        super().__init__()

    @torch.cuda.amp.autocast(enabled=False)
    def forward(self, x):
        return torch.fft.fft(torch.fft.fft(x, dim=-1), dim=-2).real
        
class FiTBlock(nn.Module):
    def __init__(self, config):
        super().__init__()
        self.fft = FFTLayer()
        self.layerNorm1 = nn.LayerNorm(
            config.embed_dim, eps=1e-12)
        self.ff = nn.Linear(
            config.embed_dim, config.dim_feedforward)
        self.dense = nn.Linear(
            config.dim_feedforward, config.embed_dim)
        self.layerNorm2 = nn.LayerNorm(
            config.embed_dim, eps=1e-12)
        self.dropout = nn.Dropout(config.dropout_rate)
        self.activation = nn.GELU()

    def forward(self, x):
        fftOut = self.fft(x)
        x = self.layerNorm1(fftOut + x)
        x = self.ff(x)
        x = self.activation(x)
        x = self.dense(x)
        x = self.dropout(x)
        x = self.layerNorm2(x + fftOut)
        return x
\end{minted}
Note:  \lstinline{PyTorch} library was used for more efficient computation of FFT on \lstinline{cuda} GPUs.
The code blocks are only a snippet of the entire much larger codebase used for this project. The entire code can be found at https://github.com/siddagra/fourier-project

A much smaller patch embedding size can be used in this methodology, while still having a fast, efficient, and lightweight model. A smaller patch embedding size allows the model to extract more robust localised features for classification.
Similar to ViT, a special CLS token is introduced which is passed through a linear projection and a $GeLU$ activation function to predict the class of the image.

Categorical-cross entropy loss is used for the multi-class classification:
$$
-\sum_{c=1}^My_{i,c}\log(p_{i,c})
$$
where $p_{i,c}$ is the predicted probability of image $i$ belonging to class $c$ and $y_{i,c}$ is a binary indicator (0 or 1) of the ground truth label: whether the image is actually of class $c$.

Results:

\begin{table}[h!]
\centering
\begin{tabular}{c c c c} 
 \hline
 Model & Accuracy & Inference Time (ms) & Params \\ [0.5ex] 
 \hline
 ViT & 93.5 & 5.67 & 86M \\ 
 FiT & 94.3 & 3.6 & 38M \\
 \hline
\end{tabular}
\caption{Comparision of Vision Transformer (ViT) and Fourier Image Transformer (FiT) on CIFAR-10 Image Classification Benchmark (32x32 Images and 10 classes)}
\label{table:1}
\end{table}

Patch size for ViT was set to 4x4 and patch size for FiT was set to 2x2 for the benchmark on CIFAR-10 dataset.

The results do not show much of a difference. Likely due to the simple dataset of 10 classes and low resolution 32x32 images. The major advantage of such scaling schemes is only apparent when testing on higher resolution images. However, training on such larger images and datasets would take a few days of compute.

\begin{figure}[!h]
    \centering
    \includegraphics[width=\linewidth/2]{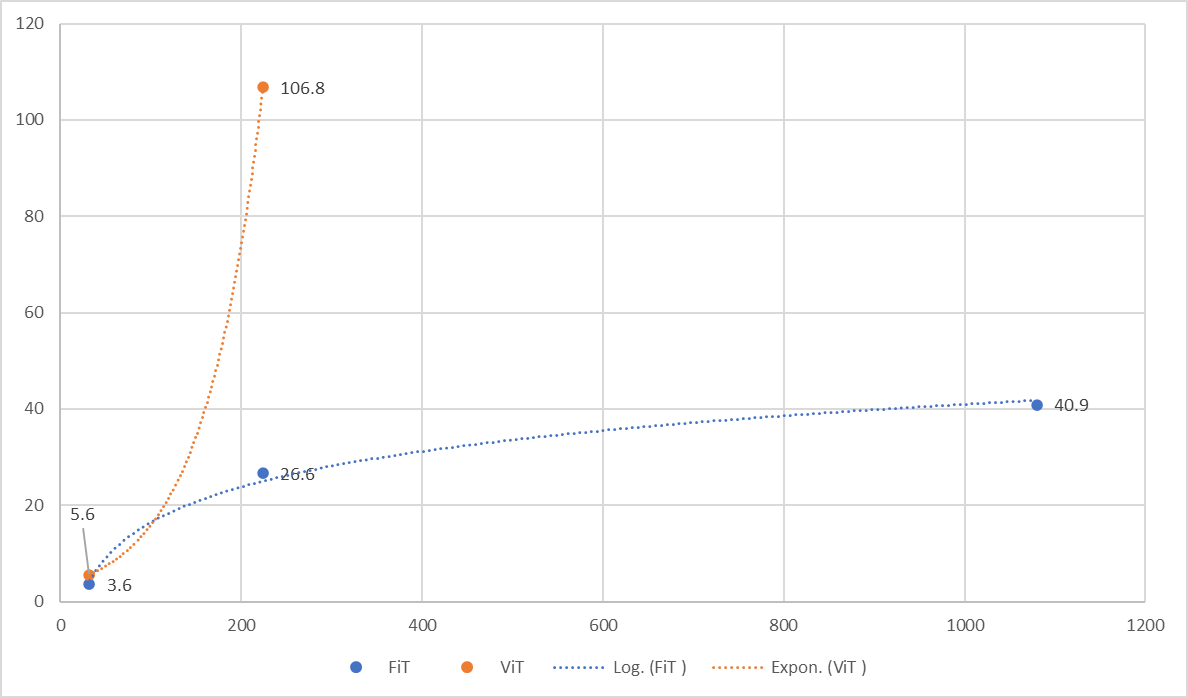}
    \caption{Comparision of ViT (orange line) and FiT (blue line) inference times on larger images}
    \label{fig:galaxy}
\end{figure}

Still, FiT achieves a higher accuracy while using much less parameters and having a faster inference speed. The parametric efficiency is attained due to the fact that unlike self-attention, token mixing using FiT requires no learnable parameters. Besides, FFT also has better scalability to larger sequences due to its $O(nlog(n))$ complexity

We can also see that FiT scales up in inference time much better than ViT:

\begin{table}[!ht]
    \centering
    \begin{tabular}{llll}
    \hline
        Image Size & Patch Size & FiT  & ViT  \\ \hline
        32x32 & 4x4 & 3.6ms & 5.6ms \\ 
        224x224 & 4x4 & 26.6ms & 106.8ms \\ 
        1080x1080 & 16x16 & 40.9ms & DNF \\ 
    \hline
    \end{tabular}
\caption{Comparision of ViT and FiT inference times on larger images}
\end{table}

\section{Scaling Up Convolutions}
State-of-the-art Convolutions Neural Networks (CNNs) have been attempting to scale up the kernel size of convolutions in order to increase performance of the networks. They have found that depth-wise separable convolutions perform better at larger kernel sizes. This is where the convolution is only applied in the spatial dimensions and not the channel dimensions. However, as kernels get larger, the efficiency of computing results using the model reducing drastically due to the quadratic complexity. For each convolution kernel of size $m\times m$ and an image of size $n\times n$, a total of $m^2\times n^2$ operations are required to compute depth-wise separable convolutions. Large neural networks use several large kernels in order to perform well on vision tasks. Thus, a solution is needed if further scalability is required. Convolution using the Fourier Transform presents one such possible solution. 

Multiplication in the frequency domain is equivalent to convolutions in the spatial domain.

Proof:
$$
\begin{aligned}
\mathcal{F}[f * g] &=\int_{-\infty}^{\infty} g(z)\int_{-\infty}^{\infty} f(x-z) e^{-2 \pi i x} d x d z \\
&=\int_{-\infty}^{\infty} g(z)\int_{-\infty}^{\infty} f(y) e^{-2 \pi i (y+z)} d y d z \\
&=\int_{-\infty}^{\infty} g(z) e^{-2 \pi i z} d z\int_{-\infty}^{\infty} f(y) e^{-2 \pi i y} d y \\
&=\mathcal{F}[f] \cdot \mathcal{F}[g]
\end{aligned}
$$

Via the Fast Fourier Transform, convolution can be computed between a kernel of size $m\times m$ and an image of size $n\times n$ with $O(n^2log(n))$ time complexity. With this, the run-time complexity of the convolution no longer depends on the kernel size, but solely on the image resolution, allowing us to scale up the convolution kernel to arbitrarily large sizes.

However, in 2D, element-wise multiplication in the fourier domain actually results in circular convolutions and not linear convolutions. The circular convolution is periodic and repeats with length $n$ (image size), whereas a linear convolution would result in an output of size ($n+(m-1)$), where $m$ is the size of the kernel. This causes the image signal to be squeezed in size, causing aliasing artifacts. The $(m-1)$ values wrap around due to the periodicity of the circular convolution and interfere with the actual image signal. To convert this into a linear convolution, we can simply first pad the image by $(m-1)$, to allow the image size to be $n+(m-1)$ itself, i.e, the period of the circular convolution. These can then be circularly shifted back to its original position and the padded values can be cropped out from the result of the convolution operation.

We first pad the input image and the kernel to same dimensions to have a linear convolution:
\begin{minted}{python}
padded_image = torch.nn.functional.pad(self.image, (self.kernel_size-1), value=0.0)  
padded_kernel = torch.nn.functional.pad(self.kernel, self.image_size + self.kernel_size - 1, value=0.0)  
\end{minted}
Next, we compute fourier transform of image and kernel using FFT:
\begin{minted}{python}
## computed in the last two dimensions to be depthwise seperable
image_ft = torch.fft.rfftn(image, dim=(-1,-2))
kernel_ft = torch.fft.rfftn(kernel, dim=(-1,-2))
\end{minted}

Convolutional Neural Networks utilise cross-correlation and not convolution (despite its name). Cross-correlation does not flip the kernel unlike convolution. Definition of cross-correlation:
$$
(f \star g)(x)=\int_{-\infty}^{\infty} f(x+z) g(z) d z=h(x)
$$

While this does not matter for the model as the kernel is learnable and the model may learn to flip the kernel if required, however, as I want this to be directly transferable to state-of-the-art CNN models, I need to implement cross-correlation instead of convolution. This can be done by taking the complex conjugate of the fourier transformed kernel (kernel\_ft):

\begin{minted}{python}
kernel_ft.imag *= -1
output_ft = image_ft * kernel_ft
\end{minted}

Finally, the inverse fourier transform of this is computed to bring the signal back into the spatial domain.
\begin{minted}{python}
output = torch.fft.irfftn(output_ft, dim=(-2,-1))
\end{minted}

We remove the padding done previously to attain our final cross-correlation output:
\begin{minted}{python}
crop_dim = (self.image_size + self.kernel_size - 1)
output = output[:, :, :crop_dim, :crop_dim] # B, C, H, W
\end{minted}

Lastly, many CNN models optionally apply a learnable bias term before returning the output:
\begin{minted}{python}
if bias:
    output += bias.view(1, -1, 1)
\end{minted}

It can be seen that the value of this operation produces similar results to the spatial convolutions under both the $\mathcal{L1}$ and $\mathcal{L2}$ norms.

\begin{figure}[htp]
    \centering
    \includegraphics[width=\linewidth/4]{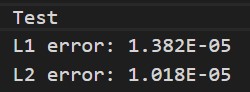}
    \caption{Comparision with native PyTorch Convolution}
    \label{fig:galaxy}
\end{figure}

\begin{figure}[htp]
    \centering
    \includegraphics[width=\linewidth]{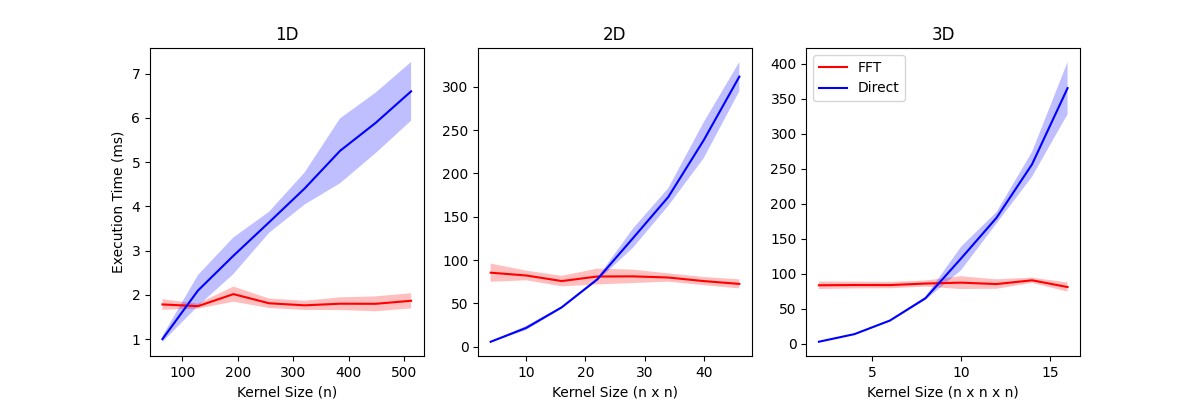}
    \caption{Scalibility of fourier convolutions\cite{fkodom_2022}}
    \label{fig:galaxy}
\end{figure}

The best part about this implementation is that it can be directly used with pre-trained state-of-the-art CNN models without any modifications to the architecture other than switching the convolutional function to call \verb|FFT_Convolution()|. Table 3 shows replacing convolutions in RepLKNet \cite{ding_zhang_zhou_han_ding_sun_2022} with fourier convolutions and the speedup acquired through it. It also demonstrates that the accuracy on ImageNet test dataset benchmark largely remained the same.

\begin{table}[ht!]
\centering
\begin{tabular}{c c c c} 
 \hline
 Model & Accuracy & Inference Time (ms) & Params \\ [0.5ex] 
 \hline
 RepLKNet-base & 83.5 & 41.2 & 79M \\
 FFT-Conv-RepLKNet-base & 83.4 & 28.7 & 79M \\
 \hline
\end{tabular}
\caption{ImageNet-1k benchmark performance (224x224 image size)}
\label{table:1}
\end{table}

\newpage
\subsection{Structured State Space and Hippo ODEs}
$$
\begin{aligned}
x^{\prime}(t) &=\boldsymbol{A} x(t)+\boldsymbol{B} u(t) \\
y(t) &=\boldsymbol{C} x(t)+\boldsymbol{D} u(t)
\end{aligned}
$$

A, B, C, D are learnable parameters of the Machine Learning model. These are used often in simpler Machine Learning models such as Hidden Markov Model (HMM), however, they can also be extended to recurrent neural networks such as GRUs and LSTMs. 

However, recent developments in this field have found that formulating such state spaces as a convolutional kernel (modified by $\boldsymbol{C}$) formed using a particular basis kernels $K_n(t)$ (controlled by $\boldsymbol{A}, \boldsymbol{B}$ ):

$$
K(t)=\sum_{k=0}^{N-1} \boldsymbol{C}_k K_k(t) \quad K_n(t)=\left(e^{t \boldsymbol{A}} \boldsymbol{B}\right)_n
$$

\begin{figure}[htp]
    \centering
    \includegraphics[width=\linewidth]{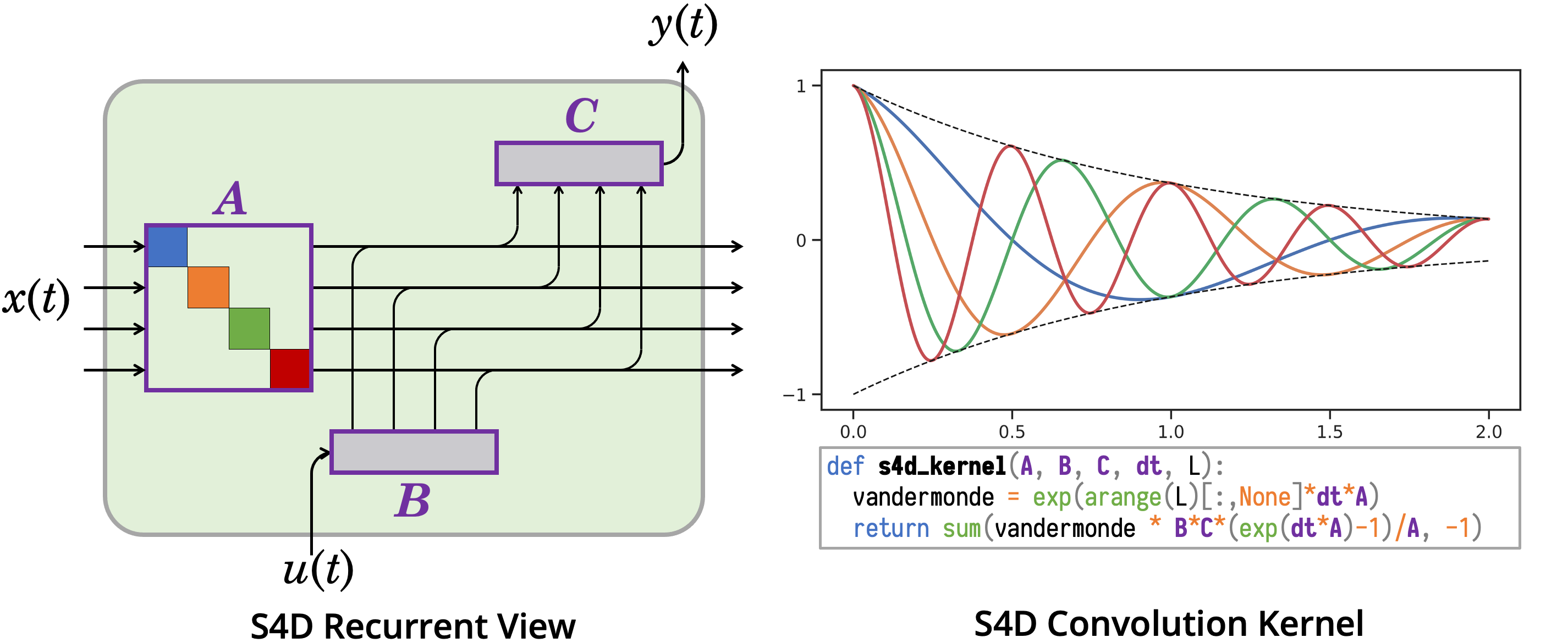}
    \caption{Overview of S4D\cite{gu_gupta_goel_ré_2022}}
    \label{fig:galaxy}
\end{figure}

If A and B are not restricted at all, they suffer from vanishing and exploding gradients, where backpropogated errors accumulate, leading to either extremely large or extremely small gradients during gradient descent/backpropogation. This leads to poor convergence and performance, and sparser learnable matrices. By controlling the matrices such that the eigenvalues are always 1, it allows the matrices to remain more stable.

Recent work \cite{hippo} has found that by formulating restricting A and B to basis functions that are known to approximate sequences well, there can be a considerable improvement in the performance of state spaces (increase in performance from ~60\% to ~98\% in MNIST-sequential benchmarks). By introducing this structure to state spaces, it improves there long term memory capacity and also gives some structure to the matrices A and B. 

The original work restricted $A$ to the class of legendre polynomials. And also introduced $lagT$ which is a legendre polynomial with a learnable exponentially decaying factor, giving less importance to long term memory for more recent events.

Structured State Spaces (S4) restrictes A and B to:
$$
A_{n k}=\left\{\begin{array}{ll}
(2 n+1)^{1 / 2}(2 k+1)^{1 / 2} & \text { if } n>k \\
n+1 & \text { if } n=k, \\
0 & \text { if } n<k
\end{array} \quad B_n=(2 n+1)^{\frac{1}{2}}\right.
$$

Furthermore, such matrices ODEs can be reformulated into convolutions:

$$
\begin{aligned}
&\begin{aligned}
x^{\prime}(t) &=\boldsymbol{A} x(t)+\boldsymbol{B} u(t) \\
y(t) &=\boldsymbol{C} x(t)
\end{aligned}\\
&\begin{aligned}
K(t) &=\boldsymbol{C} e^{t \boldsymbol{A}} \boldsymbol{B} \\
y(t) &=(K * u)(t)
\end{aligned}
\end{aligned}
$$

Thus FFT can be used to convolute the kernel $K$ (controlled by $A$ and $B$) with the sequence.

Furthermore, the time complexity of computing the kernel convolution can also be reduced to $O(N+L)$ using this structured state space, as compared to $O(N^2+L)$ in the conventional non-structured state spaces. Where $N$ is the number of parameters in $K$ and $L$ is the length of the sequence. This can be done due to the structure introduced to these matrices, which consequentially produces a Diagonal Plus Low-Rank matrix for $K$. This can be converted into 4 weighted dot products by using the Woodburry Identity. I am not going too much into detail into this as it is out of the scope of this report. More importantly, the next section will introduce methods that are more efficient and perform better, and do not require this step.

\subsection{Approximating Large Convolutional Kernels}
\cite{li_cai_zhang_chen_dey_2022} found that the reason why S4D kernels perform so well was predominantly due to two major points: 1) \textbf{efficient parameterisation} as provided by restricting the matrices $A$ and $B$ to specific basis functions, and 2) \textbf{decaying structure}, which provides a good inductive bias to long-sequence modelling as the magnitude of the value of the convolution kernel decays as it approaches the current time step, so that more weight is assigned to the most recent neighbours, because the spectrum of the power of a matrix decays exponentially.

S4 produces kernels such as this:

\begin{figure}[htp]
    \centering
    \includegraphics[width=\linewidth]{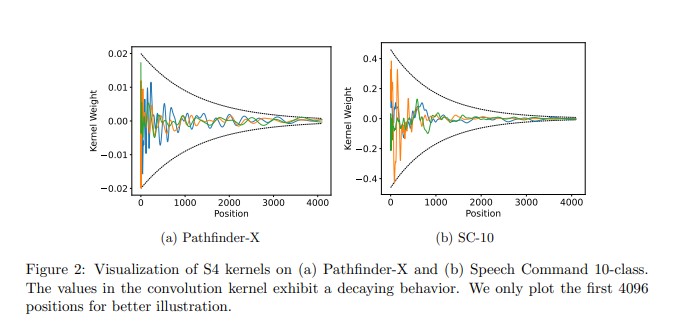}
    \caption{S4 kernel visualisations\cite{li_cai_zhang_chen_dey_2022}}
    \label{fig:galaxy}
\end{figure}

They set out to find a different, more efficient reparametirsation and found that simply resizing the kernel using bilinear interpolation to be equal to the image/sequence size works slightly better, and requires much fewer parameters. The kernel is scaled up to sequence length/image size and convolution is computed using Fast Fourier Transform and element-wise multiplication. 

This can be generalized to 2D or N-D by either taking the outerproduct of two such kernels, or by flattening the image sequence and modelling it as a larger 1D sequence. 
They match the performance of ConvNexts \cite{convnext} while using much less parameters.

Code implementation done by me:
\begin{minted}{python}
class GConv(hk.Module):
    def __init__(self, width, depth=96, bidirectional=True):
        self.width = width
        self.depth = depth
        self.bidirectional = bidirectional

    @hk.transparent
    def kernel(self, seq_length):
        scale_count = np.ceil(np.log(seq_length) / np.log(2)).astype(int)
        scales = 1 / 2 ** jnp.arange(scale_count)
        concat = []
        kernel = hk.get_parameter(
            "kernel", (self.width, self.depth), init=hki.RandomNormal()
        )
        for i, scale in enumerate(scales):
            concat.append(
                jax.image.resize(
                    kernel * scale, (self.width * 2**i, self.depth), method="bilinear"
                )
            )
        kernel = jnp.concat(concat)
        if self.bidirectional:
            kernel = ein.rearrange("(k n) d -> k n d", k=2)
            kernel = jnp.concatenate([kernel, kernel], axis=0)
        kernel = jnp.take(kernel, jnp.arange(seq_length), axis=0)
        return kernel

    def __call__(self, signal):
        seq_length = signal.shape[-2]
        k_f = jnp.fft.rfft(self.kernel(seq_length), axis=-2)
        u_f = jnp.fft.rfft(signal, axis=-2)
        y_f = k_f * u_f
        y = jnp.fft.irfft(y_f)
        b = hk.get_parameter("bias", self.depth)
        return y + b
\end{minted}
Note: I could not test this due to time and computational limitations. Training takes very long. However, I did test over a few epochs and loss is reduced to a level that is better than random guessing. i.e., $log(\verb|num_classes|)$

This convolution function is directly swappable into ConvNext, replacing the native \verb|Conv2D| function.

In this case, the original kernel is learned through backpropogation, and the decaying structure is imposed using the \verb|scales| which exponentially decays the kernel. The kernel, through backpropogation, learns to produce values that scale efficiently when resized using bilinear interpolation, and the exponential decay is introduced to assign more weightage to the nearest neighbours in the current timestep.

\bibliographystyle{unsrt}  
\bibliography{references}

\end{document}